\documentclass[letterpaper, 10 pt, conference]{ieeeconf}  

\IEEEoverridecommandlockouts                              

\usepackage{graphics} 
\usepackage{epsfig} 
\usepackage{mathptmx} 
\usepackage{times} 
\usepackage{amsmath} 
\usepackage{amssymb}  
\usepackage{graphicx}
\usepackage{cite}
\usepackage{gensymb}
\usepackage{multirow}

\title{\LARGE \bf
2D3D-MatchNet: Learning to Match Keypoints
Across 2D Image and 3D Point Cloud
}

\author{Mengdan Feng$^{1}$, Sixing Hu$^{2}$, Marcelo H Ang Jr$^{1}$ and Gim Hee Lee$^{2}$
\thanks{$^{1}$Department of Mechanical Engineering, National University of Singapore
        {\tt\small fengmengdan@u.nus.edu, mpeangh@nus.edu.sg}}%
\thanks{$^{2}$Computer Vision and Robotic Perception (CVRP) Lab, Department of Computer Science, National University of Singapore
        {\tt\small hu.sixing@u.nus.edu, gimhee.lee@nus.edu.sg}}%
}

\begin{document}

\maketitle
\thispagestyle{empty}
\pagestyle{empty}

\begin{abstract}
Large-scale point cloud generated from 3D sensors is more accurate than its image-based counterpart. However, it is seldom used in visual pose estimation due to the difficulty in obtaining 2D-3D image to point cloud correspondences. In this paper, we propose the 2D3D-MatchNet - an end-to-end deep network architecture to jointly learn the descriptors for 2D and 3D keypoint from image and point cloud, respectively. As a result, we are able to directly match and establish 2D-3D correspondences from the query image and 3D point cloud reference map for visual pose estimation. We create our \textit{Oxford 2D-3D Patches dataset} from the Oxford Robotcar dataset with the ground truth camera poses and 2D-3D image to point cloud correspondences for training and testing the deep network. Experimental results verify the feasibility of our approach. 
\end{abstract}

\section{Introduction}
\label{introduction}
Visual pose estimation refers to the problem of estimating the camera pose with respect to the coordinate frame of a given reference 3D point cloud map. It is the foundation of visual Simultaneous Localization and Mapping (vSLAM) \cite{orb-slam,lsd-slam}, and Structure-from-Motion (SfM) \cite{SfM}, which are extremely important to applications such as autonomous driving \cite{auto-vehicle} and augmented reality \cite{AR} etc. A two-step approach ~\cite{SfM} is commonly used for visual pose estimation - (1) establish 2D-3D keypoint correspondences between the 2D image and 3D reference map, and (2) apply the Perspective-n-Point (PnP) \cite{p3p} algorithm to compute the camera pose with respect to the coordinate frame of the 3D reference map with at least three 2D-3D correspondences. The 3D point cloud of the reference map is usually built from a collection of images using SfM, and the associated keypoint descriptors, e.g. SIFT \cite{SIFT}, are stored with the map to facilitate the establishment of 2D-3D correspondences in visual pose estimation. 

It is eminent that the accuracy of visual pose estimation strongly depends on the quality of the 3D reference map. Unfortunately, it is hard to ensure the quality of the 3D point cloud reconstructed from SfM, since most images are taken by close-to-market photo-sensors that are noisy. Furthermore, absolute scale of the reconstructed 3D point cloud is not directly available from SfM and has to be obtained from other sources. The need to store keypoint descriptors from multiple views of the same keypoint also increases the memory consumption of the map. In contrast, a 3D point cloud map built from Lidars has the advantages of higher accuracy and absolute scale is directly observed. Despite the advantages, Lidars are seldom used to build the 3D reference map because of the lack of a descriptor that allows direct matching of the keypoints extracted from a 2D image and 3D point cloud.  


In this paper, we propose the 2D3D-MatchNet - a novel deep network approach to jointly learn the keypoint descriptors of the 2D and 3D keypoints extracted from an image and a point cloud. We use the existing detectors from SIFT~\cite{SIFT} and ISS~\cite{iss} to extract the keypoints of the image and point cloud, respectively. Similar to most deep learning methods, an image patch is used to represent an image keypoint, and a local point cloud volume is used to represent a 3D keypoint. We propose a triplet-like deep network to concurrently learn the keypoint descriptors of a given image patch and point cloud volume such that the distance in the descriptor space is small if the 2D and 3D keypoint are a matching pair, and large otherwise. The descriptors of the keypoints from both the image and point cloud are generated through our trained network during inference. The EPnP~\cite{epnp2009} algorithm is used to compute the camera pose based on the 2D-3D correspondences. We create our~\textit{Oxford 2D-3D Patches} dataset with 432,982 pairs of matching 2D-3D keypoints based on the Oxford dataset. We conduct extensive experiments on our dataset to verify that sufficient inliers for visual pose estimation can be found based on our learned feature vectors. Pose estimation succeeds without any prior on 2D-3D correspondences. 

\textbf{Contributions}
(1) To the best of our knowledge, we are the first to propose a deep learning approach to learn the descriptors that allow direct matching of keypoints across a 2D image and 3D point cloud. (2) Our approach makes it possible for the use of Lidars to build more accurate 3D reference map for visual pose estimation. (3) We create a dataset with huge collection of 2D-3D image patch to 3D point cloud volume correspondences, which can be used to train and validate the network. 



\section{Related Work}
\label{sec:relatedwork}
\textbf{Traditional localization}  Most existing work on visual pose estimation can be classified into two categories: (1) local structure based methods and (2) global appearance based methods. 2D-3D correspondences are first established from SIFT features given the query image and 3D scene model~\cite{efficientloc, worldwideloc} for local structure based methods. Each local feature pair votes for its own pose independently, without considering other pairs in the image. Then a minimal solver algorithm, e.g.~\cite{epnp2009}, combined with RANSAC iterations, is used for robust pose estimation. Global appearance based localization methods~\cite{global-loc, ulrich2000} aggregate all local features of the query image to a global descriptor and localize the query image by matching against its nearest neighbour in current image database as a retrieval problem. 

\textbf{Learnable localization} 
Deep learning is increasingly applied to the visual pose estimation since learned features are shown to be more robust against environmental changes, e.g. lighting and weather changes, compared with methods based on hand-crafted features such as SIFT~\cite{loc-lstm}. Existing deep network models solve the localization problem from different aspects. \cite{posenet, loc-lstm} learn to regress the 6D camera pose directly based on single image. \cite{sceneforest, regforest} learn to predict pixel-wise image to scene correspondences, followed by a RANSAC-like optimization algorithm for pose estimation. \cite{dsac} proposes to localize an image by learnable probabilistic 2D-3D correspondences and iterative refinement of camera pose. However, they cannot generalize to unseen environment due to the global pose estimation mechanism. \cite{semloc} proposes to learn robust 3D point cloud descriptors by fusing semantic and geometric information for long-term localization.

\textbf{Deep similarity learning} 
Deep similarity learning is widely used to achieve the information retrieval task. Two common architectures of deep similarity learning are the Siamese network and the triplet network. The Siamese network learns the similarity relationship between a pair of inputs~\cite{Zagoruyko2015, Han2015}. Most existing work shows better results on the Triplet architecture~\cite{Hu2018, Liao2018, Vo2016, Guo2016, Schroff2015}. Liao $et.~al$~\cite{Liao2018} use the triplet network to re-identify a person's identity. Vo and Hays~\cite{Vo2016} conduct experiments on the ground-to-aerial image retrieval task using both Siamese architecture and Triplet architecture. They show that Triplet architecture performs better. The Triplet network outperforms the Siamese network because it can jointly pull the positive sample to the anchor while pushing the negative sample away. In our work, our proposed network is based on the Triplet network.


\section{Approach}
\label{sec:approach}

In this section, we outline our pipeline for visual pose estimation with a 2D query image and 3D point cloud reference map built from Lidar scans. We first introduce the overview of our pipeline in section~\ref{subsec:overview}. In section~\ref{subsec:network}, we describe our novel 2D3D-MatchNet - a deep network to jointly extract the descriptors of the 2D and 3D keypoints from an image and a point cloud. The training loss is given in section~\ref{subsec:loss}.
Finally, we discuss the pose estimation algorithm we use to compute the camera pose given at least three 2D-3D correspondences in section~\ref{subsec:pose_estimation}.


\subsection{Overview}
\label{subsec:overview}
Given a query image $I$ and the 3D point cloud reference map $\mathit{M}$ of the scene, the objective of visual pose estimation is to compute the absolute camera pose $P=[R~|~t]$ of the query image $I$ with respect to the coordinate frame of the 3D point cloud reference map $\mathit{M}$. Unlike existing visual pose methods which associate image-based descriptors, e.g. SIFT \cite{SIFT}, to each 3D point in the reference map, we propose the 2D3D-MatchNet - a deep network to jointly learn the descriptors directly from the 2D image and 3D point cloud. We first apply the SIFT detector on the query image $I$ to extract a set of 2D keypoints $U = \{u_1, ..., u_N~|~u_n \in \mathcal{R}^2\}$, and the ISS detector \cite{iss} on the 3D point cloud of the reference map $\mathit{M}$ to extract a set of 3D keypoints $V = \{v_1, ..., v_M~|~v_m \in \mathcal{R}^3\}$. Here, $N$ and $M$ are the total number of 2D and 3D keypoints extracted from the image $I$ and point cloud $\mathit{M}$, respectively. Given the set of 2D image patches centered around each 2D keypoint and 3D local point cloud volume centered around each 3D keypoint, 
our 2D3D-MatchNet learns the corresponding set of 2D and 3D descriptors denoted as $P = \{p_1, ..., p_N~|~p_n \in \mathcal{R}^{D}\}$ and $Q = \{q_1, ..., q_M~|~q_m \in \mathcal{R}^{D}\}$ for each corresponding 2D and 3D keypoint in $U$ and $V$. $D$ is the dimension of the descriptor. Furthermore, the descriptors $P$ and $Q$ learned from our network yield a much smaller similarity distance $d(p, q)$ between a matching pair of 2D-3D descriptors $(p, q)$ in comparison to the similarity distance $d(\bar{p},\bar{q})$ between a non-matching pair of 2D-3D descriptors $(\bar{p}, \bar{q})$, i.e. $d(p,q) \ll d(\bar{p},\bar{q})$, thus establishing the 2D-3D correspondences between $P$ and $Q$. Finally, 
the 2D-3D correspondences found from our 2D3D-MatchNet are used to estimate the absolute pose of the camera using a PnP algorithm. We run the PnP algorithm within RANSAC \cite{ransac} for robust estimation. 



\begin{figure}
	\centering
	\includegraphics[width=\linewidth]{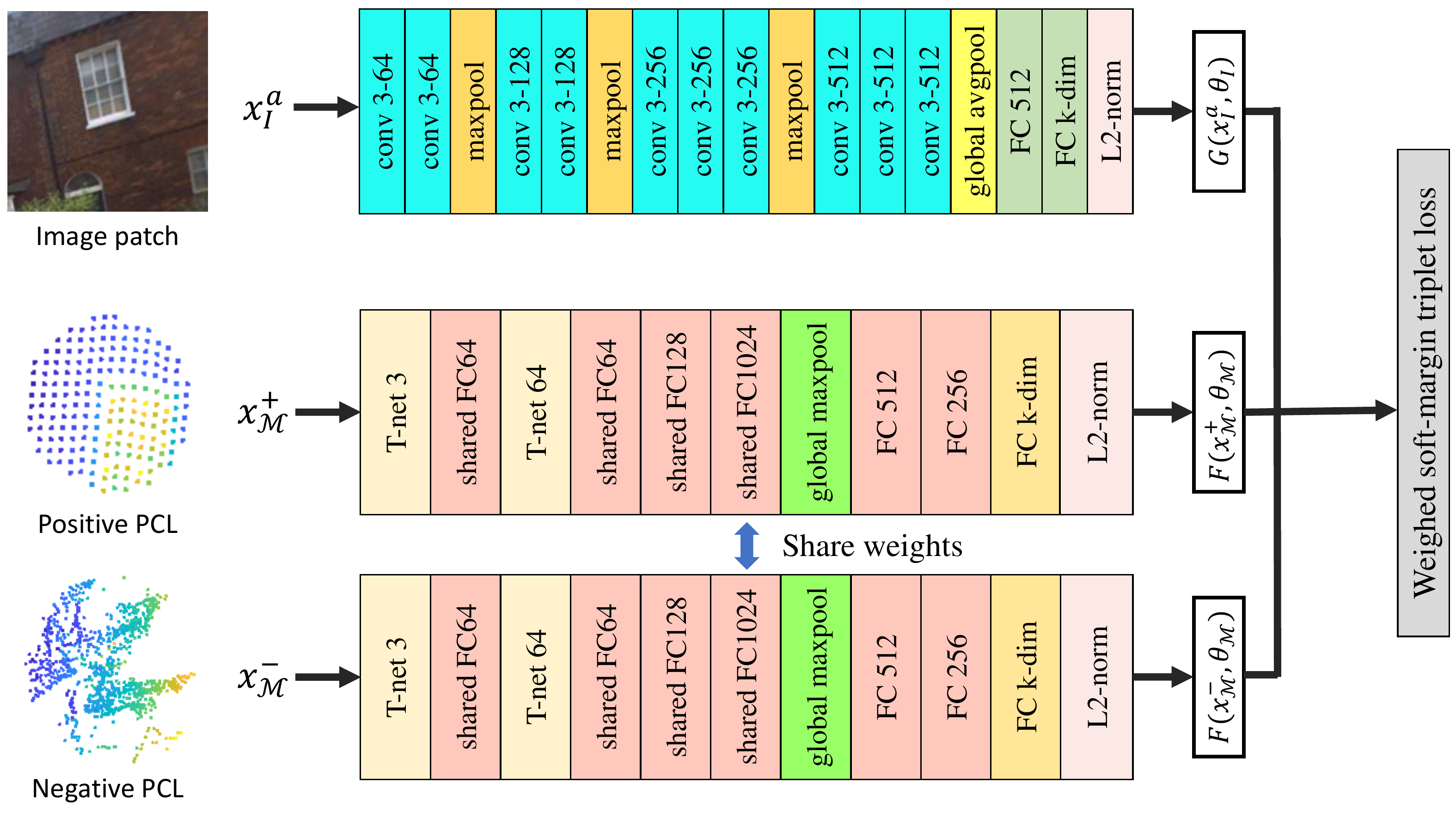}
	\vspace{-0.4cm}
	\caption {Our triplet-like 2D3DMatch-Net.}
	\vspace{-0.1cm}
	\label {fig:network}
\end{figure}

\subsection{Our 2D3D-MatchNet: Network Architecture}
\label{subsec:network}
Our 2D3D-MatchNet is a triplet-like deep network that jointly learns the similarity between a given pair of image patch and local point cloud volume. The network consists of three branches as illustrated in Fig.~\ref{fig:network}. One of the branches learns the descriptor for the 2D image keypoint and the other two branches with shared weights learn the descriptor for the 3D point cloud keypoint. The inputs to the network are (1) image patches centered on the 2D image keypoints, and (2) local volume of point cloud within a fixed radius sphere centered on the 3D keypoints. Details on keypoints extraction and the definitions of image patch and point cloud sphere are given in Sec.~\ref{subsec:training_data}. 

The image patches and local volume of point clouds are fed into the network during training as tuples of anchor image patch, and positive and negative local volume point cloud. We denote the training tuple as $\{x^a_I, x^{+}_{\mathit{M}}, x^{-}_{\mathit{M}}\}$. Given a set of training tuples, our network learns the image descriptor function $G(x_I; \theta_I):x_I \mapsto p$ that maps an input image patch $x_I$ to its descriptor $p$, and the point cloud descriptor function $F(x_{\mathit{M}}; \theta_{\mathit{M}}):x_{\mathit{M}} \mapsto q$ that maps an input local point cloud volume $x_{\mathit{M}}$ to its descriptor $q$. $\theta_I$ and $\theta_{\mathit{M}}$ are the weights of the network learned during training. More specifically:

\textbf{Image Descriptor Function} 
We design $G(x_I; \theta_I)$ as a convolutional neural network followed by several fully connected layers to extract the descriptor vector of an image patch. 
We use the well-known VGG network~\cite{VGG} as the basis of our image descriptor function network. We make use of the first four convolution block (conv1  $\sim$ conv4) of VGG16 to make the network fit better for image patch with small size. 
Global average pooling is applied on the feature maps from conv4. Compared to the max pooling layer which is widely used after convolutional layers, global average pooling
has the advantages of reducing the number of parameters and avoiding over-fitting. Two fully connected layers are concatenated at the end of the network to achieve the desired output descriptor dimension. The output descriptor vector is L2-normalized before feeding into the loss function. 

\textbf{Point Cloud Descriptor Function}
We use the state-of-art PointNet~\cite{Qi2017} as our point cloud descriptor function $F(x_{\mathit{M}}; \theta_{\mathit{M}})$ to extract the descriptor vector of a local point cloud volume. The dimension of the last fully connected layer is changed to fit our feature dimension. The softmax layer in the end is replaced by a L2-normalization. 

\textbf{Loss Function for Training}
\label{subsec:loss}
Our network is trained using the Triplet loss, so that the similarity distance $d_{pos} = d(G(x_I^a; \theta_I), F(x_{\mathit{M}}^{+}; \theta_{\mathit{M}}))$ between the matching anchor $x_I^a$ and positive $x_{\mathit{M}^{+}}$ pair is small and much lesser than the similarity distance $d_{neg} = d(G(x_I^a; \theta_I), F(x_{\mathit{M}}^{-}; \theta_{\mathit{M}}))$ between the non-matching anchor $x_I^a$ and negative $x_{\mathit{M}^{-}}$ pair, i.e. $d(G(x_I^a; \theta_I), F(x_{\mathit{M}}^{+}; \theta_{\mathit{M}})) \ll d(G(x_I^a; \theta_I), F(x_{\mathit{M}}^{-}; \theta_{\mathit{M}}))$. Specifically, Our network is trained with the weighted soft-margin triplet loss~\cite{Hu2018}: 
\begin{equation}
    \mathcal{L} = ln(1 + e^{\alpha d}),
\end{equation}
where $d = d_{pos} - d_{neg}$. We use this loss because it allows the deep network to converge faster and increase the retrieval accuracy~\cite{Hu2018}. In contrast to the basic Triplet loss~\cite{Schroff2015, Chechik2010}, it also avoids the need of selecting an optimal margin. In our experiment, we set $\alpha = 5$.
We use the Euclidean distance between two vectors as the similarity distance $d(.,.)$ in this work. During inference, we check the distance $d(G(x_I; \theta_I), F(x_{\mathit{M}}; \theta_{\mathit{M}}))$ between the descriptors of a given pair of image patch $x_I$ and local point cloud volume $x_{\mathit{M}}$. $x_I$ and $x_{\mathit{M}}$ are deemed matching pair if the distance is lesser than a threshold, and non-matching pair otherwise.  

\subsection{Pose Estimation}
\label{subsec:pose_estimation}
The pose of the camera is computed from the putative set of 2D-3D correspondences obtained from our 2D3DMatch-Net. Specifically, we obtain the 2D keypoints of the 2D query image with the SIFT detector, and the 3D keypoints of the 3D point cloud with the ISS detector. 
We compute the 2D and 3D keypoint descriptors with our network from the image patches and local point cloud volume extracted around the keypoints. The similarity distance is computed for every pair of 2D and 3D keypoints, and we find the top $K$ closest 3D point cloud keypoints for every 2D image keypoint. Finally, we apply the
EPnP algorithm~\cite{epnp2009} to estimate the camera pose with all the putative 2D-3D correspondences. The EPnP algorithm is ran within RANSAC for robust estimation to eliminate outliers.



\section{Dataset}
\label{sec:dataset}
In this section, we present the creation of our benchmark dataset -- \textit{Oxford 2D-3D Patches Dataset}. The dataset contains a total of 432,982 image patch to pointcloud pairs, which allows sufficient training and evaluation for the 2D-3D feature matching task.

\subsection{The Oxford 2D-3D Patches Dataset}
Our Oxford 2D-3D Patches dataset is created based on the Oxford RobotCar Dataset~\cite{RobotCarDataset}. The Oxford RobotCar Dataset collects data from different kinds of sensors, including cameras, Lidar and GPS/INS, for over one year. We use the images from the two (left and right) Point Grey Grasshopper2 monocular cameras, the laser scans from the front SICK LMS-151 2D Lidar, and the GPS/INS data from the NovAtel SPAN-CPT ALIGN inertial and GPS navigation system. 
Ignoring the traversals collected with poor GPS, night and rain, we get 36 traversals for over a year with sufficiently challenging lighting, weather and traffic conditions. We synchronize the images from the left and right cameras, and 2D laser scans from the Lidar with the timestamps, and get their global poses using the GPS/INS data. We remove camera and Lidar frames with small motion. 

\begin{figure}[t]
	\centering
	\includegraphics[width=0.9\linewidth]{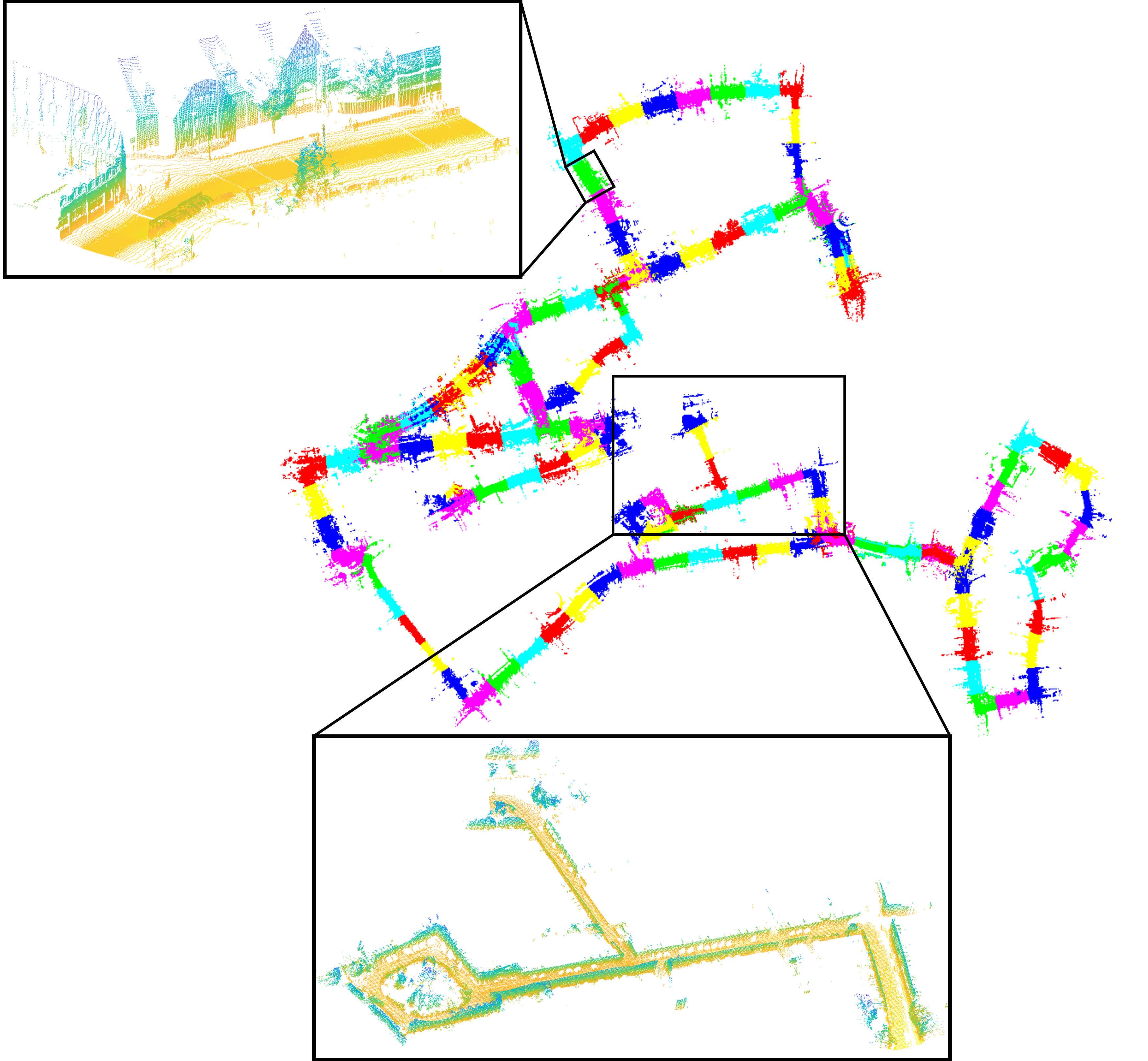}
	\vspace{-0.3cm}
	\caption {Reconstructed point cloud map from Lidar. Different colors represent different submaps. Two zooming in examples of point cloud are shown, the top one indicates a 60m submap, the bottom shows the last unseen 10\% of the path for testing.}
	\label {fig:pointcloud_map}
\end{figure}

To simplify point cloud processing and reduce the detrimental effects of GPS jumps over long distance, we split each traversal into disjoint submaps at every 60m interval. Each submap contains the corresponding sets of left and right cameras and Lidar frames. A visualization of the reconstructed point cloud map from Lidar scans is illustrated in Fig.~\ref{fig:pointcloud_map}.

\subsection{Training Data Generation}
\label{subsec:training_data}

\textbf{Keypoint Detection} 
We build a point cloud based reference map from the laser scans for every submap, where the coordinates of the first laser scan is used as the reference frame. We detect the ground plane and remove all points lying on it. This is because the flat ground plane is unlikely to contain any good 3D keypoint and descriptor. The ISS keypoint detector is applied on the remaining point cloud to extract all 3D keypoints. We apply the SIFT detector on every image to extract all 2D keypoints. 

\textbf{2D-3D Correspondences} 
To establish the 2D-3D correspondences, we project each ISS keypoint to all images within its view and find the nearest neighbour of SIFT keypoint in each image. To increase the confidence of the correspondences, we require the distance of the projected nearest neighbour to be smaller than 3 pixels and each ISS point must have at least SIFT correspondences in three different views within each submap. The ISS points and their corresponding SIFT points that satisfy these requirements are reserved for further processing.

\begin{figure}
	\centering
	\includegraphics[width=\linewidth]{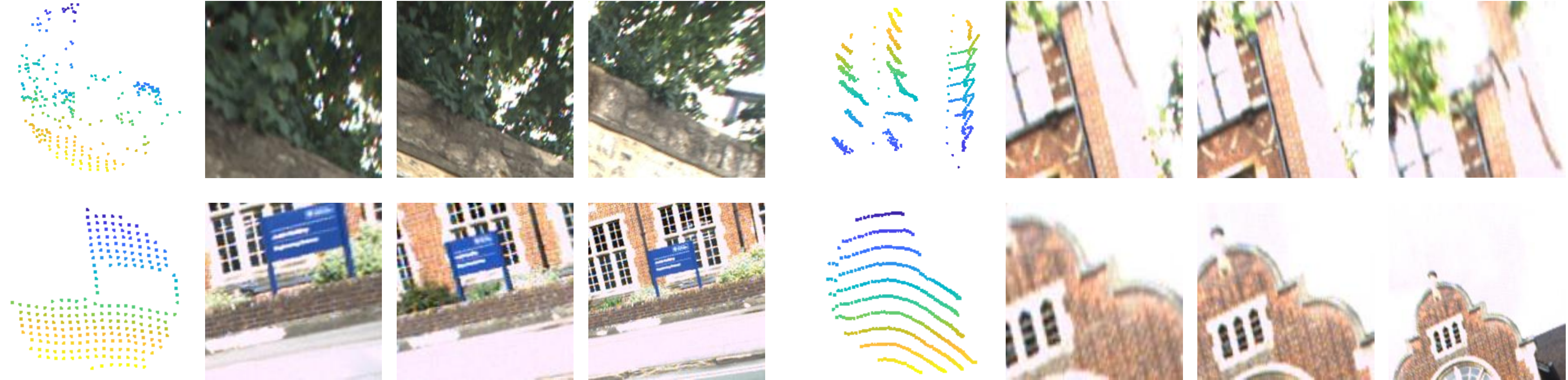}
	\vspace{-0.3cm}
	\caption {Four examples of our dataset. The first image of each example shows the ISS volume. The other three are some corresponding SIFT patches across multiple frames with different scale, viewpoint and lighting.}
	\label {fig:dataset}
\end{figure}

\textbf{ISS Volume and SIFT Patch Extraction} 
We remove all ISS keypoints that are within 4m from a selected ISS keypoint in each submap, and remove all SIFT keypoints within 32 pixels from a selected SIFT keypoint in each image. 
We find all 3D points that are within 1m radius from each selected ISS keypoint, and discard those ISS keypoints with less than 100 neighboring 3D points.
We discard SIFT keypoints with scale larger than a threshold value, since larger scale results in smaller patch size. In our experiments, we set this threshold value as 4 and the patch size at the basic scale as $256\times256$. We vary the extracted patch size with the scale of the extracted SIFT keypoints for better scale invariance.
In our experiments, we extract the ISS volume and its corresponding SIFT patch if the number of points within the ISS volume is larger than 100 and the SIFT patch is at suitable scale. We discard both the ISS volume and SIFT patch otherwise. 
Fig.~\ref{fig:dataset} shows the visualization of several examples of the local ISS point cloud volumes and their corresponding image patches with different scales, viewpoints and lightings.


\textbf{Data Pre-processing}
Before training, we rescale all the SIFT patches with different scales to the same size, i.e. 128$\times$128, and zero-center by mean subtraction. We subtract each point within each ISS point cloud volume with the associated ISS keypoint, thus achieving zero-center and unit norm sphere. Additionally, we pad the number of points to 1024 for each local volume in our experiments. 


\subsection{Testing Data Generation}
\label{subsec:testing_data}
Our objective during inference is to localize a query image based on the 2D-3D matching of the descriptors from the keypoints extracted from the image and point cloud. We test our trained network with reference submaps and images that are not used in training the network. We use the GPS/INS poses of the images as the ground truth pose for verification. The ground truth 2D-3D correspondences are computed as follows: (1) We detect all ISS keypoints from the point cloud of each submap and retain keypoints with more than 100 neighboring 3D points within 1m radius. (2) We detect SIFT keypoints on each image and extract the corresponding patches with scale smaller than the threshold value, i.e. 4 as mentioned above. (3) 
Each ISS keypoint is projected to all images within its view and the nearest SIFT keypoint 
with a distance smaller than 3 pixels is selected as the correspondence. We discard an ISS to SIFT keypoint correspondence if a nearest SIFT within 3 pixels is found in less than 3 image views.



\section{Experiments}
\label{sec:result}
In this section, we first outline the training process of our 2D3D-MatchNet that jointly learns both 2D image and 3D point cloud descriptors. Next, we describe the camera pose estimation given a query image. Then we evaluate our results and compare with different methods. Finally, we discuss and analyze the localization results of the proposed methods.   

\subsection{Network Training}
\textbf{Data splitting and evaluation metric} \hspace{0.1cm}
As mentioned in Sec.~\ref{subsec:training_data}, we split each traversal into a set of disjoint 60m submaps. We leave one full traversal for testing. For the remaining 35 traversals, we use the first 90\% submaps of each traversal for training and leave the remaining 10\% unseen for testing as in Fig.~\ref{fig:pointcloud_map}. 

We evaluate the accuracy of the estimated pose by computing the position error and the rotational error from the ground truth poses. Similar to~\cite{sattler2018benchmarking}, we define the pose precision threshold as (10m, 45\degree). We measure the percentage of query images localized within this range and report the average position error and rotation error. We choose the threshold values to satisfy the high requirements for autonomous driving.

\textbf{Network Training} \hspace{0.1cm}
Our network is implemented in Tensorflow~\cite{Tensorflow} with $2\times$ Nvidia Titan X GPUs. We train the whole network in an end-to-end manner. For each triplet input, we choose an image patch as the anchor, and its corresponding 3D point cloud volume as positive sample. The negative point cloud volume is randomly sampled from the rest of the point clouds. We initialize the image descriptor network branch with VGG model pre-trained on ImageNet~\cite{Deng2009}. Both descriptor extraction networks are optimized with Adam optimizer and the initial learning rate is $6\times10^{-5}$. The total training time is around two days.

We also explore the effect of the output feature dimension $D$ on the performance of localization. In our experiments, we train and test with different $D$ values, i.e. $D \in \{64, 128, 256\}$. The localization results from different descriptor dimensions are presented and discussed below.

\begin{table*}[h]
\centering
\centering
\small
\caption{The localization results on different traversals for over one year}
\label{tab:full_diff_run_result}
\begin{tabular}{|c|c|c|c|c|c|c|c|c|c|c|}
    \hline
    Date and Time  
                & \begin{tabular}[c]{@{}c@{}}2014.06.26 \\ 09:53:12\end{tabular}
                & \begin{tabular}[c]{@{}c@{}}2014.07.14 \\ 15:16:36\end{tabular} 
                & \begin{tabular}[c]{@{}c@{}}2015.02.03 \\ 08:45:10\end{tabular} 
                & \begin{tabular}[c]{@{}c@{}}2015.04.24 \\ 08:15:07\end{tabular} 
                & \begin{tabular}[c]{@{}c@{}}2015.06.09 \\ 15:06:29\end{tabular} 
                & \begin{tabular}[c]{@{}c@{}}2015.07.14 \\ 16:17:39\end{tabular} 
                & \begin{tabular}[c]{@{}c@{}}2015.08.13 \\ 16:02:58\end{tabular} \\ 
    \hline
    Test submaps       & 32   & 5    & 5    & 5    & 3    & 3    & 4    \\ 
    \hline
    Test frames        & 7095 & 866  & 666  & 898  & 548  & 624  & 536  \\ 
    \hline
    Average T error    & 1.41 & 1.34 & 1.88 & 1.67 & 1.68 & 1.44 & 1.71 \\ 
    \hline
    Average R error    & 6.40 & 6.62 & 7.33 & 7.24 & 7.24 & 7.17 & 7.44 \\ 
    \hline
\end{tabular}
\end{table*}

\begin{figure}
	\centering
	\includegraphics[width=0.9\linewidth]{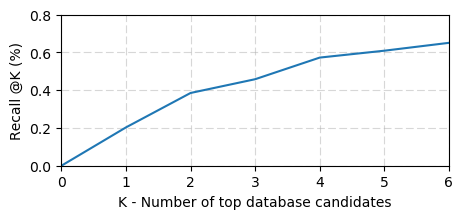}
	\vspace{-0.3cm}
	\caption {The recall in top $K$.}
	\label {fig:top_k_recall}
\end{figure}

\subsection{Results} 
We test on one full traversal and the last unseen 10\% from another six  traversals. As mentioned in \ref{subsec:testing_data}, we reconstruct the point cloud from GPS/INS data for each submap. We remove the redundant points from the ground plane. Next, we detect all the 3D keypoints and infer the corresponding descriptors from the point cloud descriptor network. All descriptors of the point cloud keypoints are stored in the database. 

\textbf{Network results} Given a query image, we extract the 2D SIFT keypoints and feed all the corresponding image patches into the image descriptor network to get the descriptors of the query image. For each image descriptor, we find its top $K$ nearest point descriptor from our database thus establishing the 2D-3D correspondences. In Fig.~\ref{fig:top_k_recall}, it shows the recall from top-1 to top-6.
The selection of $K$ can largely effect the localization results. With a larger $K$, we have more point feature candidates for each image feature. Consequently, the RANSAC algorithm is more likely to find the correct match. On the other hand, a larger $K$ unfavorably increases the number of iterations of RANSAC exponentially. Considering the trade-off, we choose $K=5$ for our experiments.

\textbf{Camera pose estimation} Finally, we solve the camera pose using the EPnP algorithm~\cite{epnp2009}. For all images in each test submap, the localization results are presented in Tab.~\ref{tab:full_diff_run_result}. The unit of T and R error are meter and degree. The first result (@2014.06.26) reports the localization results on most submaps of the full run, except for those with bad point cloud maps due to GPS inaccuracy. We denote it as $full\_test$. Others show the localization results on several different test submaps of each traversal across different times over one year. We denote them as $submap\_test$. The ratio of successfully localized frames are shown in Fig.~\ref{fig:curve}.

\begin{figure}
	\centering
	\includegraphics[width=\linewidth]{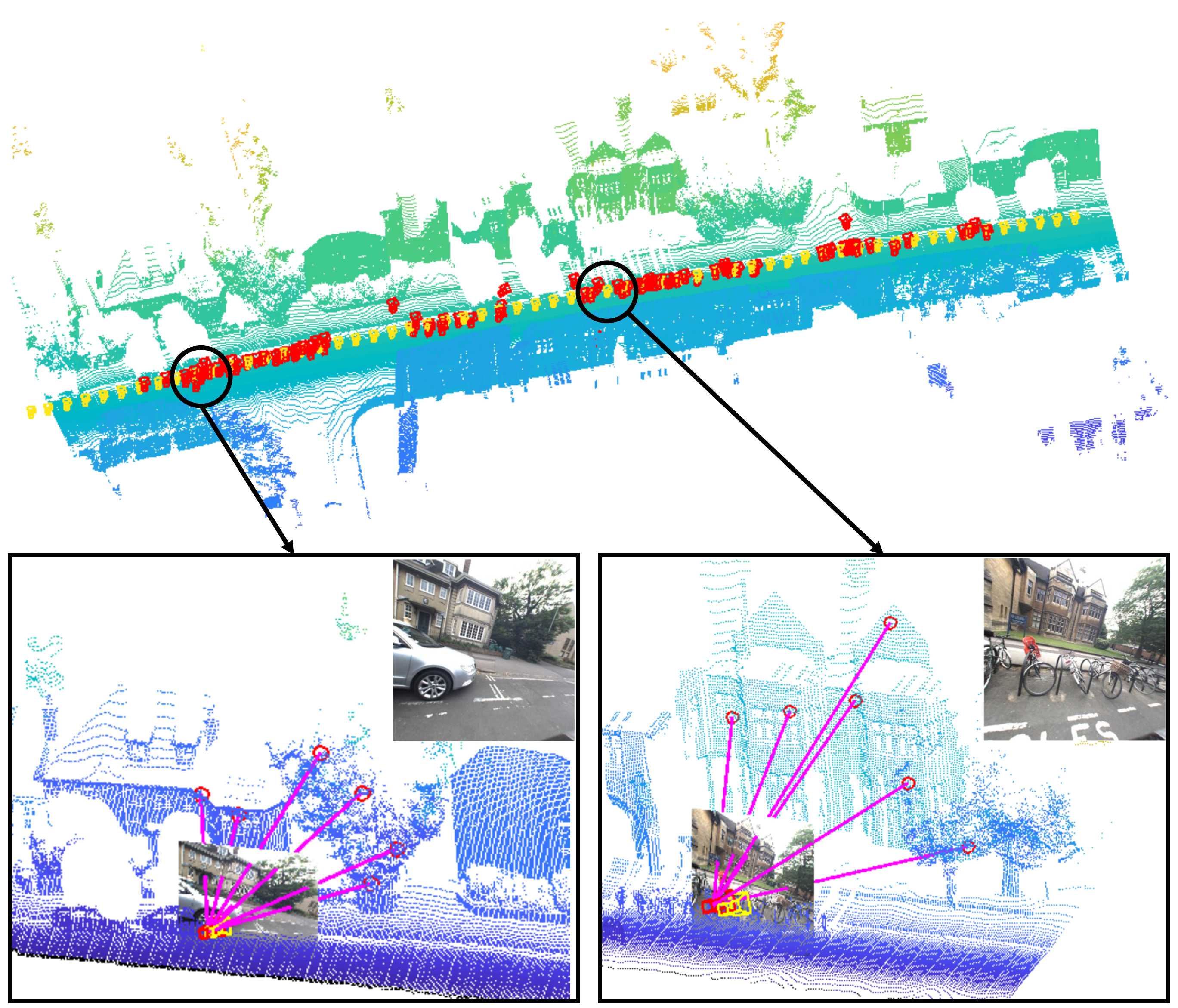}
	\vspace{-0.6cm}
	\caption {A qualitative visualization of our camera pose estimation. Red camera: our estimated camera pose. Yellow camera: the ground-truth camera pose. Purple line: some predicted 2D-3D correspondences between 2D SIFT patches and 3D ISS volumes.}
	\label {fig:submap_visual}
\end{figure}

A qualitative visualization of the localization is presented in Fig.\ref{fig:submap_visual}. In these results, the output feature dimension $D$ is set 128.

\begin{figure}
	\centering
	\includegraphics[width=\linewidth]{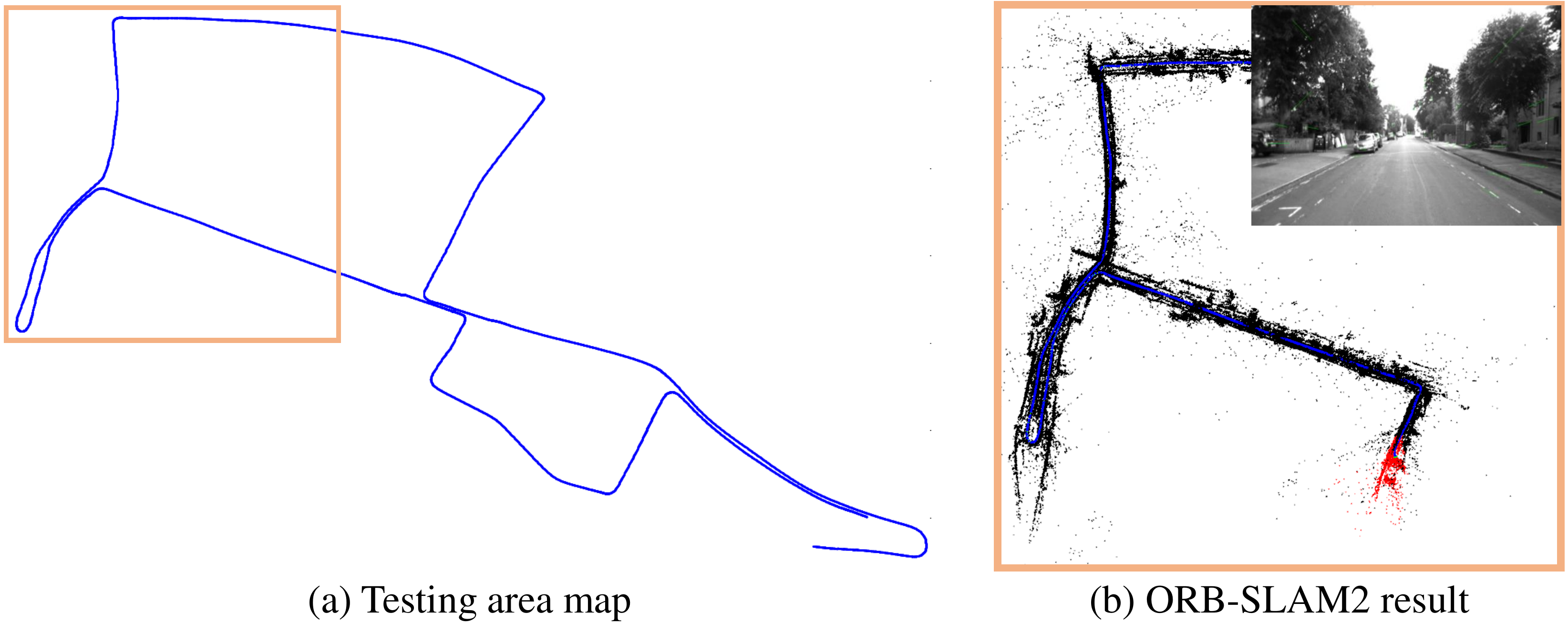}
	\vspace{-0.3cm}
	\caption {The localization result on ORB-SLAM2. ORB-SLAM2 result: The red points are the failure position. An image of the failure case is shown.}
	\label {fig:comparison}
\end{figure}

\subsection{Evaluation and Comparison}
For evaluation, we choose 8500 images collected in the overcast afternoon of July 14, 2015 with frame rate 16$Hz$. The path is around 3km with partial overlaps for loop closure detection, covering an spatial area of 845$m$ $\times$ 617$m$. We test the performance of two well-known algorithms for the task of visual localization, i.e., ORB-SLAM2~\cite{orb-slam2} (traditional method) and PoseNet~\cite{posenet} (deep learning method).

\begin{figure}
	\centering
	\includegraphics[width=\linewidth]{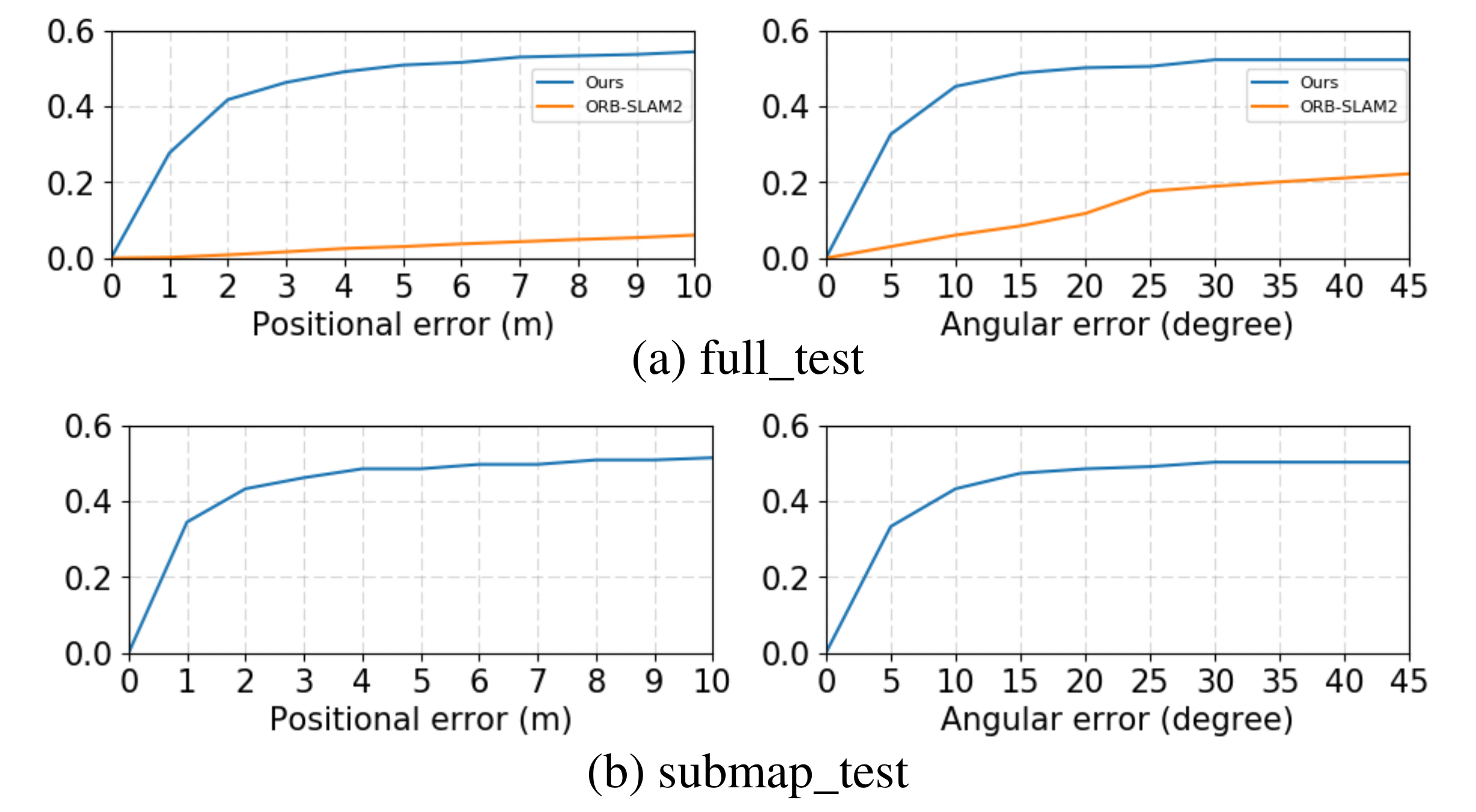}
	\vspace{-0.6cm}
	\caption {The curves of ratio of successfully localized frames w.r.t error thresholds. (a) full\_test: results of testing on the $full\_test$ with both our method and ORB-SLAM2~\cite{orb-slam2}. (b) submap\_test: average results of testing on the $submap\_test$.}
	\label {fig:curve}
\end{figure}

We use ORB-SLAM2 algorithm to build the point cloud map of the testing area. However, it cannot succeed to build the whole area. Partial mapping result is shown in Fig.~\ref{fig:comparison}-(b). The localization result using the point cloud map by ORB-SLAM2 is shown in Fig.~\ref{fig:curve}-(a). The large error is due to the inaccurate mapping. We observe that the map is unreliable when the images are captured near trees or at the turns. We train the PoseNet for visual localization. However, the localization error is huge. We argue that the PoseNet is not suitable in a large area with training data captured by cameras on a running vehicle, which do not provide rich variance on the angle and viewpoint. Furthermore, this method can not generalize to the unseen test sequences. 

The existing algorithms fail to localize images within large-scale urban environments. In the next section, we show that our algorithm can successfully localize more than 40\% of the images throughout the whole testing area.

\begin{table}[h]
\centering
\small
\caption{Localization results on different output descriptor dimension $D$ on 2015-02-13, 09:16:26}
\label{tab:diff_feature_dim}
\begin{tabular}{|c|c|c|c|c|c|}
\hline 
\begin{tabular}[c]{@{}c@{}}$D$\end{tabular} &
\begin{tabular}[c]{@{}c@{}}Test \\ frames\end{tabular} & \begin{tabular}[c]{@{}c@{}}Success \\ frames\end{tabular} & \begin{tabular}[c]{@{}c@{}}Average \\ inliers\end{tabular} & \begin{tabular}[c]{@{}c@{}}Average\\ T error\end{tabular} & \begin{tabular}[c]{@{}c@{}}Average \\ R error\end{tabular} \\ 
\hline
64    &                 & 182  & 9  & 1.18 & 6.00                   \\ 
\cline{1-1} \cline{3-6} 
128   & 625  & 187  & 10 & 1.14 & 6.10                   \\ 
\cline{1-1} \cline{3-6} 
256   &                 & 179  & 9  & 0.99 & 5.31                   \\ 
\hline
\end{tabular}
\end{table}

\subsection{Analysis and Discussion}
\textbf{Generalization} As can be seen in the Tab.~\ref{tab:full_diff_run_result}, the results of $submap\_test$ are slightly worse than the result of $full\_test$ since the testing area of $submap\_test$ is totally unseen. However, the results in unseen area are close to the results of seen area. It shows the good generalization of our proposed network.

\textbf{Output Feature Dimension $D$} We investigate the effect of the output feature dimension $D$. We test three submaps from another traversal on 2015, Feb 13, at 09:16:26. The localization result is shown in Tab.~\ref{tab:diff_feature_dim}. As we can see, the feature dimension $D$ of 128 successfully localize more images than the other two. A higher feature dimension can better represent the image and point cloud, but on the other hand, it may also cause over-fitting since our patch size and point cloud volume are small. Considering the trade-off between accuracy and inference efficiency, we choose $D$ as 128 in our experiments.

From the localization results, we show that our proposed method is able to estimate camera pose from a point cloud based reference map directly through 2D image to 3D point cloud descriptor matching using deep learning. There are two main cases where the localization is likely to fail. (1) The scene contains many trees: 3D points from trees are quite likely to be detected as key points due to their strong gradients, i.e. irregularity in shape. However, the SIFT keypoints on trees do not contain discriminative information. Consequently, wrong matches arose from patches and point clouds on trees. (2) The scene is dominated by flat building walls: buildings are always full of texture seen from image, and thus create many meaningful patches. However, points on smooth wall are less likely to be detected as keypoints. This leads to low 3D keypoints and descriptor candidates, which decreases localization performance.

\section{Conclusion}
\label{sec:conclusion}
We presented a novel method for camera pose estimation given a 3D point cloud reference map of the outdoor environment. Instead of the association of local image descriptors to points in the reference map, we proposed to jointly learn the image and point cloud descriptors directly through our deep network model, thus obtaining the 2D-3D correspondences and estimating the camera pose with the EPnP algorithm. We demonstrated that our network is able to map cross-domain inputs (i.e. image and point cloud) to a discriminative descriptor space where their similarity / dis-similarity can be easily identified. Our method achieved considerable localization results with average translation and rotation errors of 1.41m and 6.40 degree on the standard Oxford RobotCar dataset.
In future work, we aim at an end-to-end network for camera pose estimation by incorporating the hand-crafted key point selection and the RANSAC algorithm into the network. Furthermore, we will enforce temporal consistencies on multiple continuous frames to help improve the localization accuracy. 

\section{ACKNOWLEDGMENT}
This research was supported in parts by the National Research Foundation (NRF) Singapore through the Singapore-MIT Alliance for Research and Technology's (FM IRG) research programme and a Singapore MOE Tier 1 grant R-252-000-637-112. We are grateful for the support.



\clearpage


\bibliographystyle{IEEEtran}
\bibliography{icra_citations}  

\end{document}